\documentclass[conference]{IEEEtran}
\IEEEoverridecommandlockouts

\usepackage{cite}
\usepackage{amsmath,amssymb,amsfonts}
\usepackage{algorithmic}
\usepackage{graphicx}
\usepackage{multirow}
\usepackage{pifont}
\usepackage{hyperref}
\usepackage{booktabs}
\usepackage{xcolor}
\usepackage{amsmath,amssymb,amsfonts,epsfig,subfigure,bm,dsfont}
\usepackage{threeparttable}
\usepackage{algorithmic}
\usepackage[square, comma, sort&compress, numbers]{natbib}
\usepackage[linesnumbered,lined,ruled,vlined]{algorithm2e}

\makeatletter
\newcommand{\linebreakand}{%
  \end{@IEEEauthorhalign}
  \hfill\mbox{}\par
  \mbox{}\hfill\begin{@IEEEauthorhalign}
}
\makeatother

\hypersetup{
  colorlinks=true,
  linkcolor=black,
  urlcolor=black,
  citecolor=black
}

\def\BibTeX{{\rm B\kern-.05em{\sc i\kern-.025em b}\kern-.08emT\kern-.1667em\lower.7ex\hbox{E}\kern-.125emX}}
\begin{document}

\title{LAPA: Log-Domain Prediction-Driven Dynamic Sparsity Accelerator for Transformer Model\\
\thanks{Corresponding authors: Yang Hu (hu\_yang@tsinghua.edu.cn)}
}

\author{\IEEEauthorblockN{Huizheng Wang}
\IEEEauthorblockA{\textit{School of Integrated Circuits} \\
\textit{Tsinghua University}\\
Beijing, China \\
wanghz22@mails.tsinghua.edu.cn}
\and
\IEEEauthorblockN{Hongbin Wang}
\IEEEauthorblockA{\textit{School of Integrated Circuits} \\
\textit{Tsinghua University}\\
Beijing, China \\
wanghb24@mails.tsinghua.edu.cn}
\and
\IEEEauthorblockN{Shaojun Wei}
\IEEEauthorblockA{\textit{School of Integrated Circuits} \\
\textit{Tsinghua University}\\
Beijing, China \\
wsj@tsinghua.edu.cn}
\linebreakand
\IEEEauthorblockN{Yang Hu}
\IEEEauthorblockA{\textit{School of Integrated Circuits} \\
\textit{Tsinghua University}\\
Beijing, China \\
hu\_yang@tsinghua.edu.cn}
\and
\IEEEauthorblockN{Shouyi Yin}
\IEEEauthorblockA{\textit{School of Integrated Circuits} \\
\textit{Tsinghua University}\\
Beijing, China \\
\textit{Shanghai Artificial Intelligence Laboratory}\\
Shanghai, China \\
yinsy@tsinghua.edu.cn}
}

\maketitle

\begin{abstract}
Attention-based Transformers have revolutionized natural language processing (NLP) and shown strong performance in computer vision (CV) tasks. However, as the input sequence varies, the computational bottlenecks in Transformer models exhibit dynamic behavior across stages, which calls for a cross-stage sparse acceleration strategy. Unfortunately, most existing sparse Transformer approaches are single-stage based, and their sparsity prediction mechanisms lead to significant power overhead when applied across multiple stages. To this end, this paper proposes a log-domain attention prediction algorithm-architecture co-design, named LAPA. First, an asymmetric leading one computing (ALOC) scheme is designed to eliminate expensive multiplications. Next, a mixed-precision multi-round shifting accumulation (MRSA) mechanism is further proposed to mitigate the accumulation overhead. A data-feature dependent filter (DDF) strategy is designed to work in concert with the MRSA process. Finally, an elaborate accelerator is designed to translate the theoretical enhancement into practical hardware improvement. Experimental results show that LAPA achieves $3.52\times$, $3.24\times$ and $2.79\times$ higher energy efficiency than the state-of-the-art (SOTA) works Spatten, Sanger and FACT, respectively.
\end{abstract}

\begin{IEEEkeywords}
Transformers, dynamic sparsity, low complexity.
\end{IEEEkeywords}

\section{Introduction}\label{sec:introduction}
Benefiting from the powerful \emph{self-attention} mechanism \cite{vaswani2017attention}, Transformer models have achieved remarkable advances in a spectrum of tasks, spanning from text generation~\cite{devlin2018bert} to computer vision (CV)~\cite{dosovitskiy2020image}. However, such enhanced performance comes with an overwhelming computation burden. For example, although the vision Transformer G with patch 14 (ViT-G/14) \cite{zhai2022scaling}, achieves an accuracy of 90.45$\%$ on ImageNet \cite{deng2009imagenet}, showing a significant gain than non-Transformer model BiT-L (87.54\%) \cite{kolesnikov2019large}, the computation of the former is $21.3\times$ that of the latter. Such significant computational workloads necessitate the design of accelerators tailored for Transformers.  

However, a key challenge in efficiently accelerating Transformers lies in the presence of \textbf{\emph{dynamic computational bottlenecks}} \cite{kim2023full,fuad2023survey,zhao2023survey,zhou2024survey,wang2025mcbp}. Specifically, each attention block requires the computation of three matrices, i.e., Query (Q), Key (K), and Value (V), to capture pairwise dependencies among tokens within a given sequence. To enable this, these matrices are first generated through a linear projection stage, commonly referred to as \emph{QKV generation}. However, the computational complexity of QKV generation is $O(SH^2)$, while the subsequent attention computation incurs a complexity of $O(S^2H)$, where $S$ denotes the sequence length and $H$ is the hidden dimension. As a result, the dominant computation bottleneck varies with the input characteristics. For tasks involving long sequences \cite{taylong} or high-resolution images \cite{rombach2022high}, attention operations can account for $48.3\%$ to $90.4\%$ of the total computation. In contrast, for short-sequence tasks \cite{rajpurkar2016squad}, the attention module contributes less than $20\%$, with the linear projection becoming the primary computation bottleneck. Therefore, there is a pressing need for a versatile accelerator capable of efficiently handling both attention and linear projection computations.

\begin{figure}[t]
\centering
\includegraphics[width=0.85\linewidth]{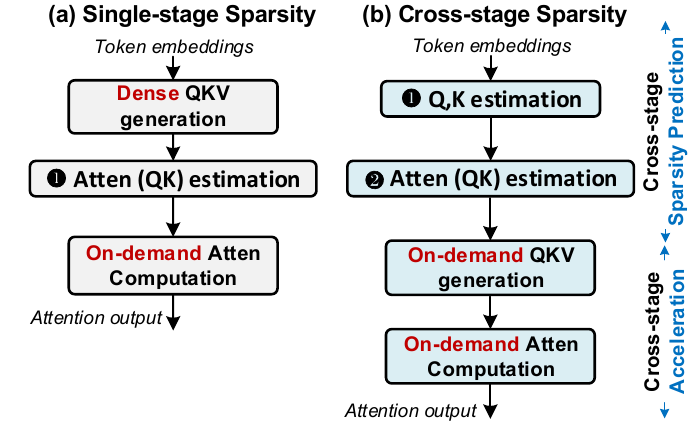}\vspace{-2mm}
\caption{Traditional single-stage sparsity and proposed cross-stage sparsity.}
\label{fig:cross-stage}\vspace{-2mm}
\end{figure}

Unfortunately, most existing efforts \cite{wang2024sofa,song2024tsacc,zhao2024hardware,wang2023cta,hong2022dfx,park2024token,wang2025beta} on Transformer sparsity focus exclusively on accelerating the attention, with limited consideration for the linear projection. To identify runtime attention sparsity, early approaches often relied on affine approximations. For instance, A$^3$~\cite{ham20203} employs progressive approximation, while ELSA~\cite{ham2021elsa} adopts hash-based techniques. However, these methods typically involve iterative processing, leading to significant latency overhead. To mitigate this, more recent works such as DOTA \cite{qu2022dota}, Energon~\cite{zhou2022energon}, and Sanger~\cite{lu2021sanger} adopt half-MSB-based computation (e.g., 4\,bit). Nonetheless, these approaches suffer from substantial power consumption during prediction. In addition, methods like SpARC \cite{cho2024sparc} and CLAT \cite{lee2025clat} simplify attention by clustering Queries. However, they rely on prior calibration, which restricts their applicability. Moreover,  they cannot be extended to linear projection scenarios. While FACT \cite{qin2023fact} realizes the challenges of the dynamic computation bottleneck and attempts to address them, its symmetric leading-one scheme introduces significant estimation inaccuracies.

To this end, this paper proposes a Cross-Stage Sparsity (CSS) strategy. As depicted in Fig. \ref{fig:cross-stage}, unlike traditional single-stage sparsity prediction, CSS first speculates the $\mathbf{\hat{Q}}$ and $\mathbf{\hat{K}}$ matrices. Based on these speculative representations, it then predicts the attention ($\mathbf{\hat{Q}}\times \mathbf{\hat{K}}$), and generates a sparse attention mask. Finally, only a small subset of $\mathbf{Q}$, $\mathbf{K}$, and $\mathbf{V}$ is generated on demand according to the mask. This approach enables the pruning of both the linear projection and attention computation associated with unimportant QKV elements.

Despite its promise, leveraging the CSS strategy imposes several significant challenges: \textbf{First, there is a lack of a low-power prediction paradigm well-suited for cross-stage processing}. While prior works on attention sparsity have explored low-power designs, such as hash-based approximation \cite{ham2021elsa} and half-MSB speculation \cite{lu2021sanger,wang2021spatten}, these approaches are limited to single-stage optimization, specifically the attention acceleration. As a result, they fail to exploit critical optimization opportunities across stages, rendering their prediction paradigms ineffective in cross-stage scenarios. \textbf{Second, there is a lack of a low-power token selection strategy tailored for cross-stage processing.} As cross-stage sparsity reduces the cost of formal computation, token selection emerges as a main contributor to power consumption. Our characterization shows that existing methods, such as top-$k$ sorting \cite{wang2021spatten}, can account for up to $30\%$ of the total power overhead in cross-stage settings. This underscores the necessity for a more efficient and low-power token selection technique.

\begin{figure}[t]
\centering
\includegraphics[width=0.95\linewidth]{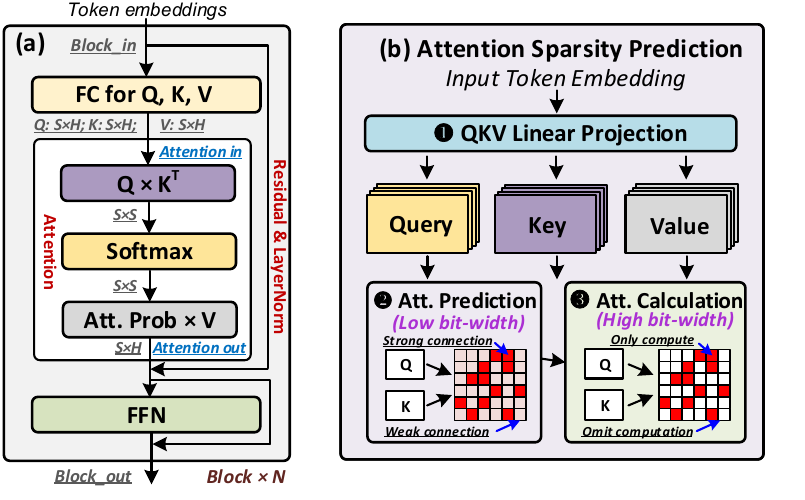}\vspace{-3mm}
\caption{The illustration for Transformer architecture and attention prediction.}
\label{fig:Transformer}\vspace{-5mm}
\end{figure}

This paper proposes LAPA, a novel software-hardware co-design to accelerate both linear projection and attention in a cross-stage manner. Our contributions include:

\textbf{1) An efficient asymmetric leading one computing (ALOC) scheme to eliminate costly multiplications, while preserving acceptable cross-stage estimation accuracy.} Moreover, the pre-known nature of weights during inference is leveraged to further reduce leading-one conversion overhead.

\textbf{2) A log-domain based multi-round shift accumulation (MRSA) scheme to progressively select the important token-pairs for decreasing overhead.} To avoid accumulation becoming a bottleneck, MRSA employs a progressive multi-round accumulation strategy combined with early-termination pruning, effectively reducing the overall accumulation overhead required for estimation.

\textbf{3) Data-feature dependent filtering (DDF) mechanism}. Leveraging the argmax property of softmax, DDF employs a simple threshold generation strategy to enable efficient pruning in MRSA, thereby reducing the estimation overhead.

\textbf{4)  Dedicated accelerator for practical acceleration}. To translate the theoretical improvement into practical performance gains, we design a customized accelerator, LAPA, to support the proposed strategies. LAPA achieves $3.52\times$, $3.24\times$, and $2.79\times$ gains in energy efficiency over state-of-the-art (SOTA) Transformer accelerators, Spatten, Sanger and FACT.

\section{Background and Motivation}\label{sec:Pre}
\subsection{Transformer Architecture}\label{subsec:Transformer}
A typical Transformer architecture consists of multiple stacked blocks, as shown in Fig.~\ref{fig:Transformer} (a). Initially, the token embedding matrix $\mathbf{X}$ is linearly projected into the query ($\mathbf{Q}$), key ($\mathbf{K}$) and value ($\mathbf{V}$) as defined in Eq.~\eqref{eq:QKV}. Then, the $\mathbf{Q}$, $\mathbf{K}$ and $\mathbf{V}$ matrices are processed by the self-attention mechanism as shown in Eq.~\eqref{eq:Attention}, where $\mathbf{A}$ denotes the attention score matrix. Further, the attention output is computed by multiplying the $\mathbf{S}$ with  $\mathbf{V}$. Notably, both the QK and SV computations have a complexity of $O(S^2H)$. The resulting intermediate results are passed through an FFN with two fully connected layers, yielding the final outputs.
\begin{equation}
\mathbf{Q},\mathbf{K},\mathbf{V}=\mathbf{XW_Q}, \mathbf{XW_K}, \mathbf{XW_V}.
\label{eq:QKV}
\end{equation}
\begin{equation}
\mathbf{S}={\rm softmax}\left({\mathbf{QK}^T}/{\sqrt{d_k}}\right), {\rm where}~\mathbf{A}=\mathbf{QK}^T.
\label{eq:Attention}
\end{equation}

\subsection{Dynamic Attention Sparsity}\label{subsec:Attention_Sparsity}
The quadratic complexity of self-attention arises from its fully-connected attention graph, where all Q-K pairs are computed. However, not all connections contribute equally to the final output~\cite{fuad2023survey}. Prior work~\cite{qu2022dota} has shown that sparse attention graphs, by excluding weak connections, can achieve performance comparable to full attention. Therefore, leveraging input-dependent dynamic sparsity presents a promising approach to reduce the computational burden of self-attention. 

Fig.~\ref{fig:Transformer} (b) illustrates the traditional workflow for attention acceleration by leveraging dynamic sparsity. First, the Q, K, V matrices are computed in a dense manner \ding{182}. Next, in the prediction stage \ding{183}, approximation schemes (e.g., 4-bit MSB) are utilized to speculate the attention distribution and generate a mask identifying the important Q–K pairs. Finally, in the formal calculation stage \ding{184},
high-precision operations (e.g., 16bit) are applied selectively to these critical Q–K pairs based on the predicted mask. However, such single-stage sparsity enables acceleration only for the attention computation, but fails to optimize the QKV generation process.

\begin{figure}[t]
\centering
\includegraphics[width=\linewidth]{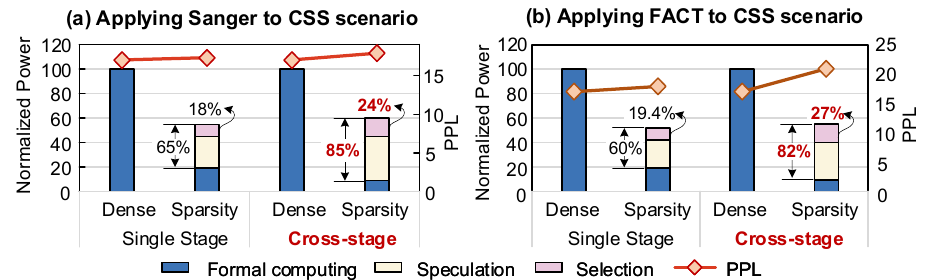}\vspace{-3mm}
\caption{Power breakdown for dense attention, single-stage sparsity attention, cross-stage sparsity attention on Llama 7B with Wikitext dataset, with TSMC 28nm CMOS. Directly applying (a) Sanger's (b) FACT's, prediction strategies to cross-stage sparsity prediction. }
\label{fig:Motivation}\vspace{-4mm}
\end{figure}

\subsection{Motivation}\label{subsec:motivation}
As discussed in Section \ref{sec:introduction}, variations in sequence length can cause the computational bottleneck in Transformers to shift between the attention mechanism and the linear projection layers. Therefore, a versatile Transformer accelerator must be capable of efficiently accelerating both components. We propose the cross-stage sparsity (CSS) strategy, which extends traditional attention sparsity to the QKV linear projections, thereby enabling sparsity acceleration across both stages. However, through a careful re-examination of existing attention sparsity techniques \cite{lu2021sanger,qin2023fact,wang2021spatten,qu2022dota}, we observe that naively applying their prediction paradigms to the cross-stage setting results in either substantial prediction power overhead or significant accuracy degradation.

Sanger \cite{lu2021sanger} and FACT \cite{qin2023fact} are SOTA and representative accelerators targeting dynamic attention sparsity. Sanger employs 4-bit MSB prediction combined with static threshold comparison, while FACT utilizes log-domain shifting followed by top-$k$ sorting. Fig. \ref{fig:Motivation} evaluates the power consumption and perplexity (PPL, lower is better) when applying the speculation strategies of Sanger and FACT to cross-stage sparsity prediction. As revealed by Fig. \ref{fig:Motivation}(a), under single-stage sparsity, Sanger's strategy successfully reduces formal computing to approximately $20\%$ of that in the dense case, while maintaining comparable PPL. In this setting, the prediction overhead, comprising speculation, i.e., $\hat{\mathbf{Q}}\times\mathbf{\hat{K}}$ and token selection, accounts for around $65\%$ of the total power. However, when applied to the cross-stage scenario, further reduction in formal computing leads to the prediction overhead rising to $85\%$, with token selection alone contributing approximately $24\%$.

On the other hand, when applying the FACT approach to cross-stage sparsity prediction, the logarithmic-domain shifting method effectively reduces speculation overhead. However, due to its larger approximation error, it leads to a noticeable degradation in PPL, approximately $20\%$ higher than the dense baseline. To make things worse, it is observed that prediction overhead accounts for as much as $82\%$ of the total power consumption, with token selection alone contributing $27\%$. 

The significant power overhead presents a major obstacle to the effective implementation of cross-stage sparsity. Therefore, enabling efficient cross-stage sparsity prediction necessitates a low-power approximation paradigm, coupled with an equally energy-efficient token selection strategy.

\begin{figure}[t]
\centering
\includegraphics[width=\linewidth]{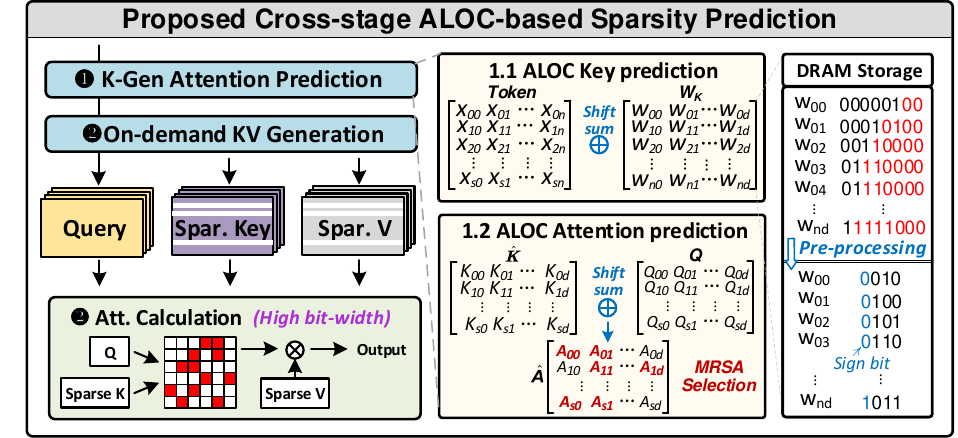}\vspace{-2.5mm}
\caption{Illustration of the ALOC-based Cross-stage sparsity (CSS) prediction.}
\label{fig:DLOC-based Cross-stage-Prediction}\vspace{-2mm}
\end{figure}

\section{Algorithm Optimizations of LAPA}\label{sec:LAPA}
To address the severe power overhead challenges posed by supporting the CSS strategy, we propose three key techniques. First, ALOC is carefully designed to exploit the known-weight property in cross-stage scenarios, effectively reducing the overhead of leading-one conversion while maintaining high prediction accuracy. Secondly, MRSA adopts a multi-round filtering mechanism to progressively identify important tokens, thereby reducing power consumption. Finally, a simple yet effective thresholding scheme is introduced to further enhance the efficiency of MRSA.

\subsection{Proposed Low-complexity ALOC Scheme}\label{subsec:ALOC}
We design an asymmetric leading one computing
(ALOC) scheme, where \emph{asymmetric} refers to the transformation of only one operand into the logarithmic domain using a leading-one encoder (LOE) to extract its leading-one (LO) position. Based on this LO value, the original multiplication is approximated by low-power shift operations applied to the other operand, thereby eliminating the need for costly multipliers. Specifically, an INT-type number $y$ can be expressed as in Eq. \eqref{eq:LoD1}:
\begin{equation}
y=Sign\cdot M\cdot 2^{LO},
\label{eq:LoD1}
\end{equation}

\begin{figure}[t]
\centering
\includegraphics[width=\linewidth]{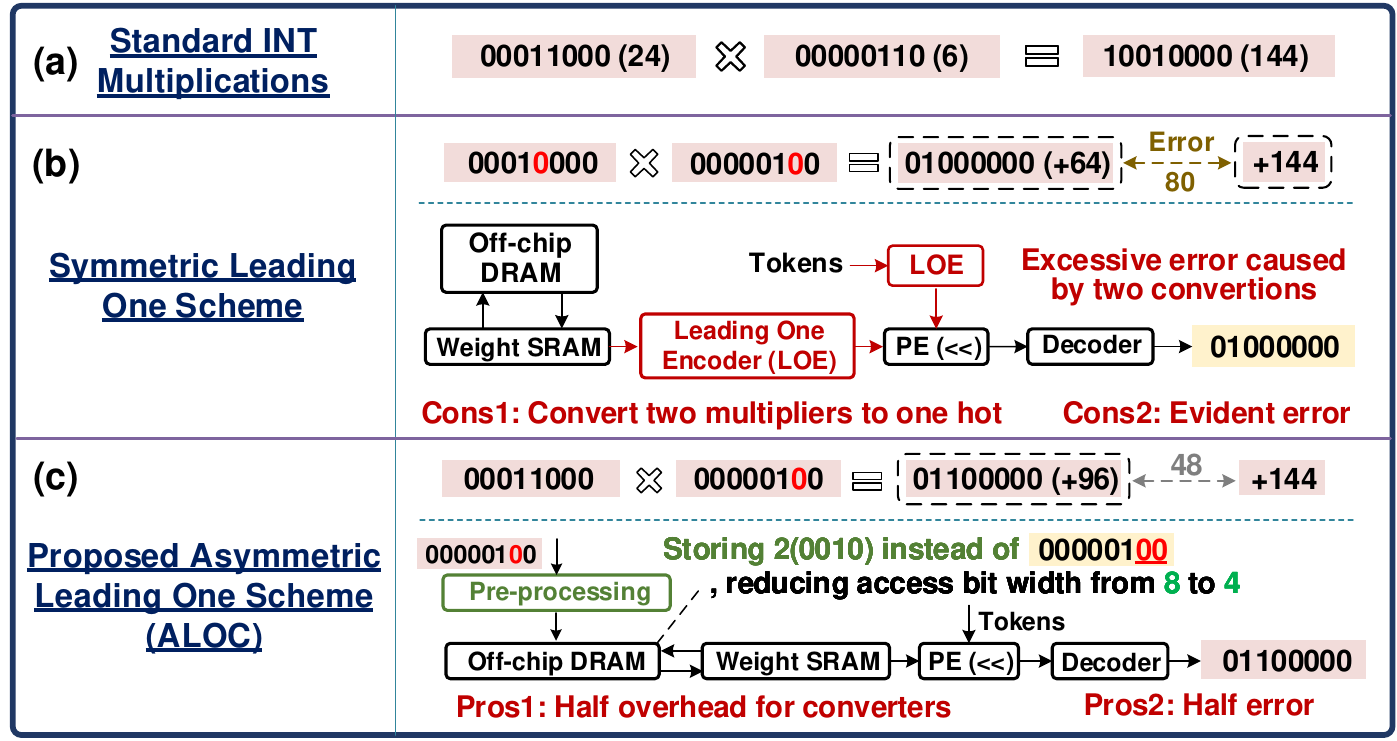}\vspace{-2mm}
\caption{Comparison of the proposed ALOC with the vanilla symmetric leading one scheme used in \cite{qin2023fact}.}
\label{fig:LoD}\vspace{-2mm}
\end{figure}

\noindent where $M$ denotes the mantissa within $[1,2)$, $LO$ denotes the leading-one position of $y$ within $[0, W-1)$, and $W$ denotes the bit-width. The corresponding multiplication is derived in Eq. \eqref{eq:LoD3} and approximated as Eq. \eqref{eq:LoD4}. Since the bit width $W$ is fixed, we can directly operate $LO_y$ on $x$, to estimate the magnitude for the product of two numbers. Therefore, by incorporating shifting and the sign bit, the multiplication result can be predicted.
\begin{subequations}
\begin{align}
&x\times y=\texttt{XOR}\left(S_x,S_y\right) \cdot \left( M_x \cdot 2^{{LO}_x}\right) \cdot \left( M_y \cdot 2^{{LO}_y}\right)\label{eq:LoD3}\\
&~~~~~~~\approx\texttt{XOR}\left(S_x,S_y\right)\cdot M_x \cdot 2^{LO_x+LO_y}\label{eq:LoD4}
\end{align}
\end{subequations}

Based on the ALOC scheme, the workflow of CSS is depicted in Fig. \ref{fig:DLOC-based Cross-stage-Prediction}. As the weights are pre-known and fixed during inference, we pre-convert the $W_K$ and $W_Q$ into LO format and store them. Then, in the Query and Key Prediction phase (1.1), no LOE is required, as the weights $\mathbf{W}_K$ and $\mathbf{W}_Q$ have been converted into LO format. In the subsequent \emph{Attention Prediction phase} (1.2), to avoid amplifying approximation error, we convert $\mathbf{\hat{Q}}$ into log domain instead of $\mathbf{\hat{K}}$, then perform shifting and sum operations. As depicted in Fig. \ref{fig:LoD}, compared to the vanilla symmetric leading one strategy used in FACT \cite{qin2023fact}, the proposed ALOC offers three key advantages: \textbf{a) Lower conversion overhead}. This is because ALOC performs the leading-one conversion on only one multiplier, while the conventional approach needs to process both. \textbf{b) Higher accuracy}. This is because the error of the leading-one scheme originates from the loss of information carried by the bits following the most significant ‘1’. Since ALOC performs the leading-one conversion on only one multiplier, it introduces less error and achieves higher precision. \textbf{c) Reduced memory access}. The reduction in memory access stems from the fact that ALOC only requires loading a 4-bit LO value that has been offline converted, whereas the vanilla symmetric leading one scheme loads the full 8-bit operand.

However, since the shifted results require sign detection and conversion into two's complement form for subsequent accumulation, the naive implementation of ALOC introduces significant bit-flipping overhead, which may even offset the benefits of the ALOC scheme. To efficiently support ALOC, we introduce two specialized types of processing elements (PEs) and a \emph{sign pre-assign} mechanism. The architectural details are discussed in Section \ref{sec:LAPA_accelerator}.

\subsection{Multi-round Shift Accumulation
(MRSA)}\label{subsec:DDF}

The ALOC strategy reduces the cost of multiplications by leveraging low-cost shift operations. However, this optimization shifts the computational bottleneck to the addition tree, which incurs significant overhead. While directly applying low-bit quantization is a feasible solution, it is a coarse-grained approach that can significantly degrade model accuracy.

\begin{figure}[t]
\centering
\includegraphics[width=\linewidth]{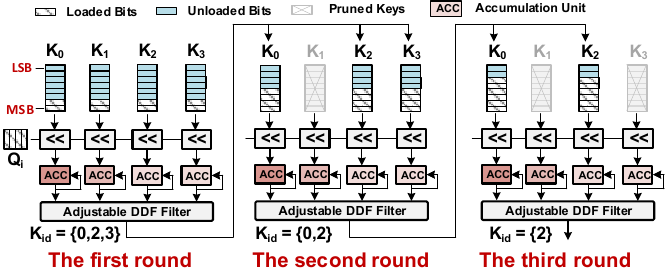}\vspace{-2mm}
\caption{Illustration for the proposed MRSA strategy with three filtering rounds.}
\label{fig:MRSA}\vspace{-2mm}
\end{figure}

 To address this limitation, we propose the mixed-precision MRSA strategy. Considering execution efficiency, MRSA is applied during the attention estimation stage. As illustrated in Fig.~\ref{fig:MRSA}, for each query $\mathbf{\hat{Q}}_i$ (in LO format), we identify significant Keys through multiple filtering rounds. Instead of employing the traditional one-shot top-$k$ pruning \cite{wang2021spatten} with low precision tensors, MRSA iteratively refines the selection via repeated filtering. Its workflow is exemplified in Fig.~\ref{fig:MRSA}. 

Initially, we utilize partial bits of Keys (e.g., high-order 2bit) to speculate $\mathbf{QK}^T$ and select pairs with relatively large scores. It is worth noting that although the low-bit speculation is not highly accurate, it remains effective in identifying significantly different values, such as the top $50\%$. In the subsequent rounds, MRSA conducts incremental filtering based on the selected pairs from the previous round. As the remaining pairs tend to have closer scores, it uses higher precision (e.g., an additional 2 bits) to further compute the dot product of the remaining Q-K pairs. This filtering process is carried out iteratively until a predefined number of rounds is reached. After the last round, the ultimately selected keys in each row are utilized for performing on-demand QKV generation and high-precision sparse attention.

\subsection{Adjustable Data-feature Dependent Filtering (DDF)} 
\begin{figure}[t]
\centering
\includegraphics[width=0.99\linewidth]{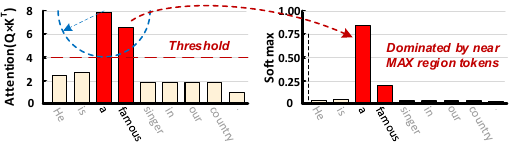}\vspace{-4mm}
\caption{Illustration for the proposed DDF strategy.}
\label{fig:DDF}
\end{figure}

The MRSA employs mixed precision for multiple rounds of filtering, making it crucial to reduce the overhead of each comparison in each round. Instead of directly sorting all the scores, we introduce the concept of a spherical radius to pick up the important scores. This is based on the following observations. As Fig.~\ref{fig:DDF} shows, the \texttt{softmax} would exponentially exaggerate the difference among the values of the inputs. Therefore, the output of the \texttt{softmax} operation is primarily determined by the input's maximum value and its neighboring values. That is to say, it is sufficient to selectively retain values that are close to the maximum.  To this end, we design an efficient data-feature-dependent filtering mechanism, which employs a max-value-centric threshold. The threshold is formally defined in Eq.~\eqref{eq:Threshold}: 
\begin{equation}
\phi_i^r=\max\left(\mathbf{A}_i^r\right)-\eta^r\times \left(\max \left(\mathbf{A}_i^r\right)-\min\left(\mathbf{A}_i^r\right)\right), \eta^r\in[0,1],
\label{eq:Threshold}
\end{equation}

\noindent where $\mathbf{A}_i^r$ denotes the attention scores of the $i$th row. To flexibly balance accuracy and pruning ratio for different application demands, we introduce a parameter $\eta^r$ to control the threshold, whose value range is $[0, 1]$. By adjusting $\eta^r$, the threshold transits from $\max \left(\mathbf{A}_i^r\right)$ to $\min\left(\mathbf{A}_i^r\right)$. Thus,
we can control the pruning ratio in each round from $0\%$ to
$100\%$.

\subsection{Overall of LAPA's CSS Mechanism}

Alg. \ref{alg:LAPA} summarizes the LAPA's CSS prediction mechanism. Given an input token sequence $\mathbf{X}$ and projection weights $\mathbf{W}_Q$ and $\mathbf{W}_K$, LAPA begins with offline preprocessing of $\mathbf{W}_Q$ and $\mathbf{W}_K$, converting them into the leading-one representation (line 2). During inference, approximate $\hat{\mathbf{Q}}$ and $\hat{\mathbf{K}}$ are generated via the ALOC strategy (line 3-4). The resulting $\mathbf{\hat{Q}}$ is then further transformed into the leading-one format (line 5). Finally, the MRSA and DDF mechanisms are applied iteratively to compute the final mask matrix (line 6-10), which captures the dynamic sparsity pattern of the attention mechanism.

\begin{algorithm}[t]\small
\caption{\texttt{Proposed CSS Mechanism}.}
\label{alg:LAPA}
$\textbf{Input:}~\mathbf{x}_i\in\mathbb{R}^{1\times H}, \mathbf{W}_Q, \mathbf{W}_K\in\mathbb{R}^{H\times H}, R$\;
$\textbf{Preprocessing:}\,\mathbf{E}^{\mathbf{W}_Q}\gets \texttt{LOE}(\mathbf{W}_Q)$;~$\mathbf{E}^{\mathbf{W}_K}\gets \texttt{LOE}(\mathbf{W}_K)$\;
$\hat{\mathbf{Q}}_{ij}\gets\sum_k\left( \texttt{Sgn}\left( X_{i,k}, W_{k,j}^Q\right) \cdot X_{i,k}<< E_{k,j}^{W_Q}\right)$\;
$\hat{\mathbf{K}}_{ij}\gets\sum_k \left(\texttt{Sgn}\left( X_{i,k}, W_{k,j}^K\right) \cdot X_{i,k}<< E_{k,j}^{W_K}\right)$\;
$\mathbf{E}^{Q}\gets \texttt{LOE}(\mathbf{\hat{Q}})$\;
\For{$r\gets 0$ {\rm to} $R$}{
$\mathbf{\hat{K}}^r\gets {\rm gather} \left(\mathbf{K}^r_{\rm id}, \mathbf{\hat{K}}\right)$\;
$\hat{\mathbf{A}}^r_{ij}\gets\sum_k \texttt{Sgn}\left( \hat{Q}_{i,k}, (\hat{K}^r_{k,j})^T\right)\cdot (\hat{K}^r_{k,j})^T  << E^{Q}_{i,k}$\;
Estimate the $\phi_i^{r}$ using Eq.~\eqref{eq:Threshold}\;
$\mathbf{K}^r_{\rm id}\gets\{k\vert \hat{A}_{i,k}^r > \phi_{i}^r \}$\;
}
$\textbf{Output:}~\mathbf{K}^R_{\rm id}$.~\tcp{Mask matrices after R rounds}
\end{algorithm}

\section{LAPA Accelerator}\label{sec:LAPA_accelerator}
\subsection{Overall Architecture}
\begin{figure}[t]
\centering
\includegraphics[width=0.97\linewidth]{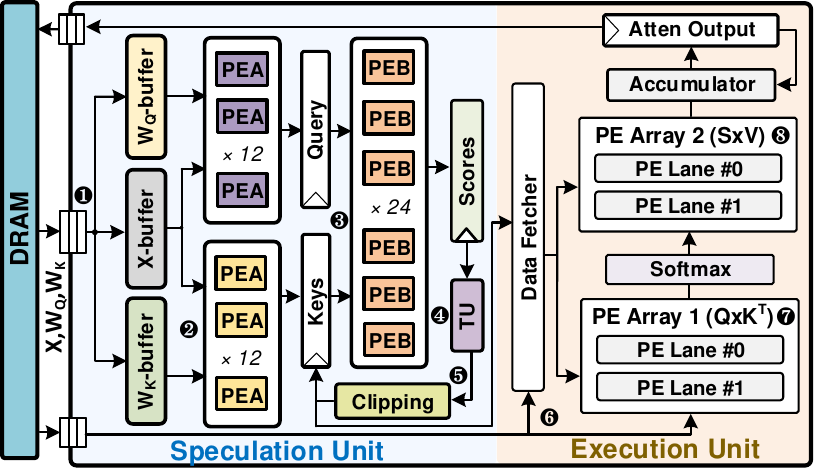}\vspace{-2mm}
\caption{High-level block diagram for LAPA accelerator.}
\label{fig:Overall_architecture}\vspace{-4mm}
\end{figure}

Fig.~\ref{fig:Overall_architecture} illustrates the overall architecture of the LAPA accelerator, which consists of a speculation unit and an execution unit. These units jointly operate to support cross-stage sparsity prediction and on-demand QKV generation.

\textbf{Overall dataflow}: The overall processing workflow is head by head. When a head's processing starts, the weight matrices $\mathbf{W}_Q$ and $\mathbf{W}_K$ are fetched from external memory in the leading one format, while the input $\mathbf{x}$ is read in the standard $8$-bit format (\ding{182}). Then, $24$ processing element A (PEA) individually computes $\mathbf{Q}=\mathbf{xW}_Q$ and $\mathbf{K}=\mathbf{xW}_K$ (\ding{183}). As depicted in Fig.~\ref{fig:Overall_architecture}, each PEA computes $64$ multiplications each time and adds up the results to obtain an inner-product result. Meanwhile, the $\mathbf{K}$ values generated by previous iterations are fetched to the K buffer. Next, the 24 PEBs calculate the approximated dot-production attention $\mathbf{\hat{A}}_i^r=\mathbf{Q}_i\mathbf{K}^T$ with multiple rounds (\ding{184}). The Threshold Updating (TU) module first provides a dynamic threshold $\phi_i^r$ (\ding{185}), which is then used by the Clipping module to identify the indices of significant keys ($K_{id}$) through comparison (\ding{186}). Based on the indices $K_{id}$ generated from the speculation unit, the Data Fetcher in the execution unit selectively retrieves the important tokens from DRAM (\ding{187}). The PE array 1 then on-demand computes the $\mathbf{K}$ and $\mathbf{V}$, as well as the corresponding attention scores (\ding{188}). These attention scores are subsequently normalized through the softmax module, and the PE array 2 performs the matrix multiplication with $\mathbf{V}$ to produce the final output (\ding{189}).

\subsection{Architecture Details}\label{subsec:Architecture_details}

Based on our experiments, for MRSA, two filtering rounds performed by the TU module achieve the optimal balance between accuracy and computation cost, and are thus adopted in our design. In this case, the first round performs coarse-grained filtering with INT4 tensors, followed by fine-grained filtering with INT8 tensors on the selected keys. This mixed-precision approach requires both INT4 and INT8 arithmetic units. To improve processing efficiency and conserve on-chip resources, we propose the result-reusable mix-precision inner production unit (IPU).

\begin{figure}[t]
\centering
\includegraphics[width=0.9\linewidth]{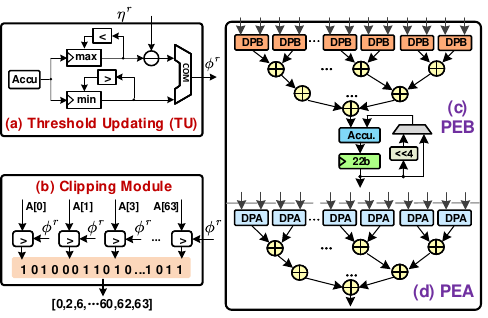}\vspace{-2mm}
\caption{Architecture components: (a) Threshold Updating module. (b) Clipping module. (c) PEB. (d) PEA.}
\label{fig:PE}\vspace{-2mm}
\end{figure}

The IPU contains multiple PEBs operating concurrently. To support both INT4 and INT8 processing, the PEB architecture is elaborated as depicted in Fig.~\ref{fig:PE} (c). This architecture not only facilitates mixed-precision processing but also diminishes computational complexity by reusing results. Each PEB equips 64 DPBs and each DPB is responsible for a 8\,bit$\times$$4$\,bit multiplication. Specifically, $\mathbf{K}^T[7:4]<<$LOE$(\mathbf{Q}_i)$ is first calculated and served as round-1 attention, then $\mathbf{K}^T[3:0]<<$LOE$(\mathbf{Q}_i)$ is computed and added with the left-shifted round-1 attention to produce round-2 attention.

\begin{figure}[t]
\centering
\includegraphics[width=0.96\linewidth]{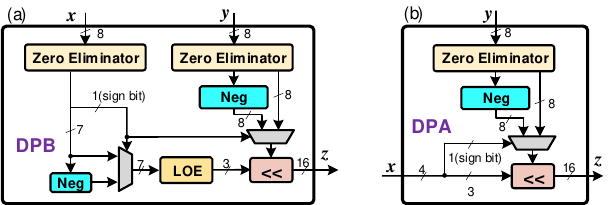}\vspace{-2mm}
\caption{The ALOC-based dot product (DP) hardware architecture.}
\label{fig:DP_architecture}\vspace{-3mm}
\end{figure}

\emph{Sign pre-assign strategy}: As discussed in Section \ref{subsec:ALOC}, naively implementing ALOC on hardware will incur severe bit-flipping overhead. To alleviate this, we propose a sign pre-assign strategy, with two types of PE support, as shown in Fig. \ref{fig:PE} (c)(d), and detailed dot-product hardware shown in Fig. \ref{fig:DP_architecture} (a)(b). Before shifting operations, the strategy involves using the sign bit of `$y$' to determine whether
to shift `$x$' itself or its opposite number. By identifying the sign of results in advance, this avoids the need for sign conversion on the product, as the product
typically owns a larger bit width, incurring higher bit-flipping overhead.

\section{Evaluation}\label{sec:Experiments}
\subsection{Experimental Setup}
\textbf{Algorithm Evaluation Setup}. We evaluate the LAPA's mechanism on various LLMs for text generation: GPT2-XL \cite{radford2019language}, OPT1B3 \cite{zhang2022opt}, Bloom1B7 \cite{le2022bloom}, Qwen7B \cite{bai2023qwen} and Llama2-13B \cite{touvron2023llama2}. To assess the impact of our methods on model performance, we measure perplexity (PPL) on the Wikitext-2 \cite{merity2016pointer} and Dolly \cite{conover2023free} datasets, where lower perplexity values indicate better performance. All evaluations are conducted with Pytorch and the HuggingFace library.

\textbf{Hardware Evaluation Details}. We implement LAPA at the RTL level and synthesize it using Synopsys Design Compiler under a 28nm CMOS technology to estimate the area and power consumption of the logic components. The performance speedup is evaluated using an in-house cycle-level simulator, with HBM2 configured as the main memory. Power, area, and read/write bandwidth are further estimated using CACTI. To evaluate the cycle count and energy consumption of off-chip memory accesses, we employ Ramulator, driven by trace files generated from RTL-level simulations.

\textbf{Design Configurations}. We evaluate the model accuracy and pruning ratio by adjusting the parameter $\eta^r$ (as Eq.\eqref{eq:Threshold}). For each round, $\eta^r$ is adjusted from $0.2$ to $0.8$ with a step of $0.1$. We then apply the successive halving method to accelerate this process. We evaluate two configurations: LAPA conservative (LAPA-C) and LAPA aggressive (LAPA-A). LAPA-C refers to settings with minimal performance degradation of at most $0.5\%$ accuracy loss, while LAPA-A is a configuration trading a slight accuracy degradation for hardware benefits, allowing for on average $2\%$ accuracy loss.

\subsection{Algorithm Performance}
\begin{figure}[t]
\centering
\includegraphics[width=\linewidth]{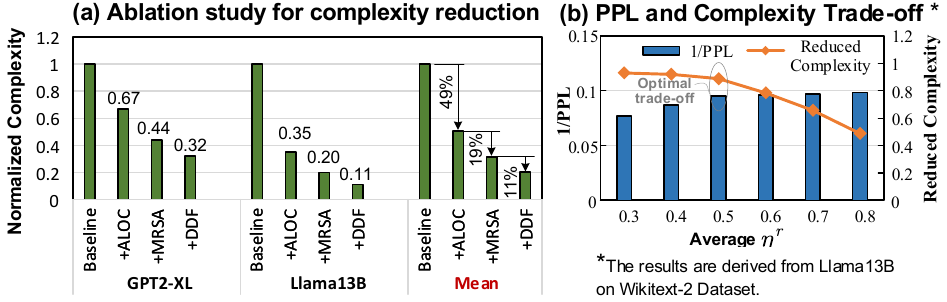}\vspace{-2mm}
\caption{(a) Complexity reduction for the ALOC, MRSA and DDF. (b) Trade-off between PPL and complexity.}
\label{fig:Ablation_study}\vspace{-2mm}
\end{figure}

We first conduct ablation experiments to evaluate the low-complexity advantages of ALOC, MRSA, and DDF against a baseline scheme, which uses vanilla leading-one operations and one-shot top-$k$ sorting for attention sparsity prediction. All operation complexities are normalized using an arithmetic complexity model \cite{brent2010modern}. As depicted in Fig. \ref{fig:Ablation_study} (a), ALOC reduces complexity by $49\%$ on average compared to the baseline, primarily due to its multiplier-free computation and half-conversion feature. MRSA achieves a $19\%$ reduction, owing to its fine-grained multi-round filtering mechanism that enables early termination of redundant operations. DDF achieves on average $11\%$ reduction, befitting from its sorting-free mechanism.

Fig. \ref{fig:Ablation_study} (b) explores the impact of various threshold parameters $\eta^r$ on model PPL and reduced complexity. Overall, a smaller $\eta^r$ implies more pruning, which helps improve reduced complexity but lowers 1/PPL. However, we observe that when $\eta^r$ drops below $0.5$, the increase in reduced complexity becomes marginal, while 1/PPL decreases dramatically. This is because overly aggressive token pruning, which hurts certain critical tokens, thereby severely degrading the 1/PPL and hindering further complexity reduction. Therefore, to balance the trade-off, we prioritize setting $\eta^r$ near $0.5$.

\begin{figure}[t]
\centering
\includegraphics[width=\linewidth]{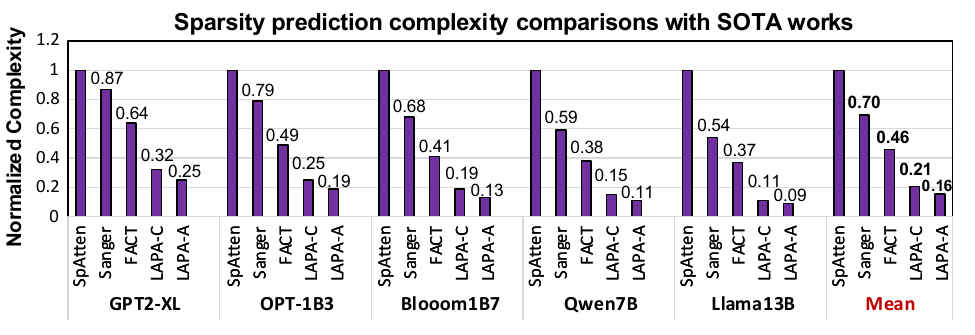}\vspace{-2mm}
\caption{Comparison of complexity reduction with SOTA works.}
\label{fig:Complexity_comparison}\vspace{-2mm}
\end{figure}

Fig. \ref{fig:Complexity_comparison} compares the sparsity prediction complexity of LAPA-C, LAPA-A, Spatten, Sanger, and FACT under various LLM models. To ensure a fair comparison, we constrain the performance degradation of each work to within $2\%$, except for LAPA-C with $0.5\%$ loss. Considering that Spatten employs multiplicative operations for attention prediction and relies on a computationally expensive top-$k$ sorting mechanism to select key tokens, it incurs the highest computational complexity. Therefore, we use it as the baseline. In contrast, Sanger utilizes a one-shot threshold-based comparison instead of sorting, achieving an average reduction of $30\%$ in computational complexity. Although FACT also adopts a log-domain multiplication-free prediction paradigm, it incurs significant conversion overhead due to the need to convert both operands, thereby limiting its complexity reduction to only $54\%$. In comparison, our proposed LAPA mechanism consistently achieves the greatest computation reduction across all models, averaging $79\%$. With slight accuracy loss, LAPA-A further achieves $84\%$ complexity reduction. 

\subsection{Hardware Evaluation}\vspace{-1mm}
Table \ref{tab:core_hardware_result} shows the area and power breakdown of the proposed LAPA accelerator. It has a total area of $3.208$ mm$^2$ and power of $474$ mW with an operating frequency of 1GHz, achieving an effective throughput of $2892$ GOPS. Although the speculation unit accounts for approximately $52\%$ of the overall area and $46\%$ of the power consumption, it effectively predicts the dynamic sparsity hidden in the self-attention, leading to an average reduction of around $79\%$ in the total computation. This, in turn, significantly reduces the area and power overheads of the execution unit. By incorporating a speculation unit, the LAPA accelerator to strike an effective trade-off, leading to significant throughput improvement.

\subsection{Comparisons with SOTA accelerators}
\begin{figure}[t]
\centering
\includegraphics[width=\linewidth]{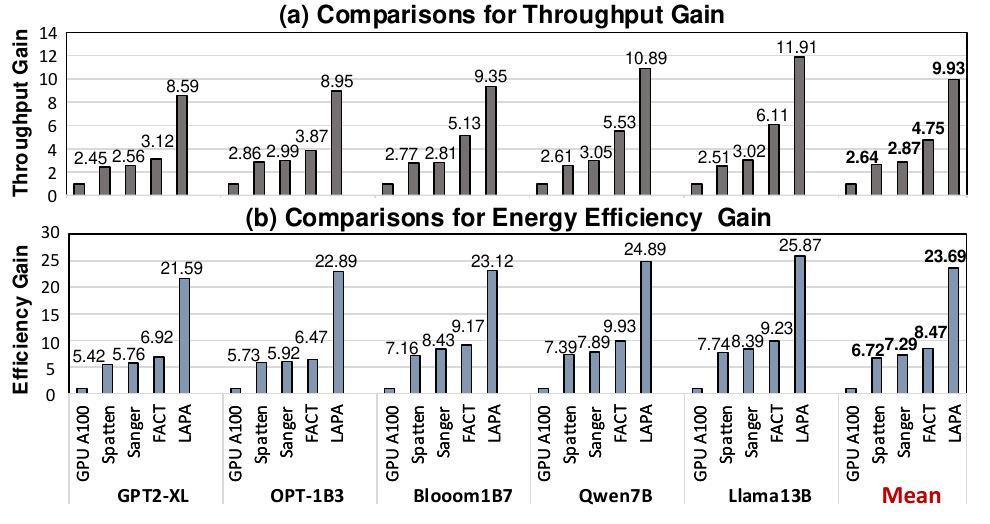}\vspace{-2mm}
\caption{Normalized attention (a) speed up and (b) energy efficiency gain.}
\label{fig:Efficiency_gain}
\end{figure}

Fig. \ref{fig:Efficiency_gain} compares the throughput and energy efficiency gains of LAPA against SOTA attention accelerators. The baseline is the Nvidia A100, which supports up to $50\%$ structured sparsity. LAPA significantly outperforms all other designs, benefiting from its hardware-friendly ALOC strategy and fine-grained multi-round filter strategy, which eliminates a large number of redundant operations with minimal overhead. On average, LAPA achieves throughput improvements of $9.93\times$, $3.76\times$, $3.45\times$, $2.1\times$ over A100 GPU, Spatten, Sanger and FACT, respectively.  In terms of energy efficiency, LAPA achieves average gains of $23.69\times$, $3.52\times$, $3.24\times$ and $2.79\times$ over the same baselines. The gain in energy efficiency primarily stems from LAPA's adoption of the ALOC scheme, which eliminates the need for costly multiplication operations.

\begin{table}[t]
\renewcommand{\arraystretch}{0.99}
\caption{Area and Power Breakdown for LAPA (Core Part) at $1$GHZ.}\vspace{-5mm}
\begin{center}
\scriptsize
\begin{tabular}{l|ccc}
\specialrule{0.12em}{0.5pt}{0.5pt}
\textbf{Units} & \textbf{Parameters} & \!\!\textbf{Area[mm$^2$]}\!\! & ~~ \textbf{Power[mW]} \\
\hline
\multirow{3}{*}{Speculation Unit} & 24 64-input 8-bit PEAs & $0.283$ & $56.54$  \\
& 24 64-input 8-bit PEBs & $0.356$ & $88.26$  \\
 & 1 TU unit, 1 Clipping unit\! &  $0.175$ & $34.05$  \\
& $512$kB Buffer &  $0.874$ & $41.87$   \\
\hline
\multirow{4}{*}{Execution Unit}  & 4 128-input 16-bit PE lines & $0.456$ & $106.54$  \\
 & 4 Softmax Modules & $0.394$  & $48.42$   \\
 & 1 accumulator & $0.108$ & $56.54$  \\
 & 256kB K/V Buffer &  $0.489$ &  $26.43$  \\
\hline
Others & - &  $0.073$ & $15.76$  \\
\hline
\textbf{Total} &  \multicolumn{3}{c}{\!\!TSMC $28$nm: Area=$3.208$mm$^2$, Power=$474.41$mW} \\
\specialrule{0.12em}{0.1pt}{0.1pt}
\end{tabular}
\end{center}
\label{tab:core_hardware_result}
\end{table}

\section{Conclusion}\label{sec:Conclu}
In this paper, we propose LAPA, a versatile Transformer accelerator that employs asymmetric log-domain multiplication-free computing, multi-round shift accumulation, and data-feature dependent filtering, to jointly realize cross-stage sparsity for both QKV linear and attention acceleration.

\section*{Acknowledgments}
This work was supported in part by the National Science and Technology Major Project under Grant 2022ZD0115200; in part by the NSFC under Grant 62125403, Grant 92464302, Grant U24B20164 and Grant 92164301; in part by Shanghai Municipal Science and Technology Major Project; in part by the Natural Science Foundation of Jiangsu Province Basic Research Program under Grant BK20243042; in part by the Beijing National Research Center for Information Science and Technology; in part by the Northern IC Technology Innovation Center (Beijing) Co., Ltd under Grant QYJS20232801B; and in part by the Beijing Advanced Innovation Center for Integrated Circuits.

\bibliographystyle{IEEEtran}
\footnotesize
\bibliography{IEEEabrv,main}

\end{document}